# Global Road Damage Detection: State-of-the-art Solutions

Deeksha Arya[*,1,2], Hiroya Maeda[2], Sanjay Kumar Ghosh[1,3], Durga Toshniwal[1,4], Hiroshi Omata[2], Takehiro Kashiyama[2], Yoshihide Sekimoto[2]

[1]Centre for Transportation Systems (CTRANS), Indian Institute of Technology Roorkee, 247667, India
[2]Institute of Industrial Science, University of Tokyo, 4-6-1 Komaba, Tokyo, Japan
[3]Department of Civil Engineering, Indian Institute of Technology Roorkee, 247667, India
[4]Department of Computer Science and Engineering, Indian Institute of Technology Roorkee, 247667, India

*Abstract.* This paper summarizes the Global Road Damage Detection Challenge (GRDDC), a Big Data Cup organized as a part of the IEEE International Conference on Big Data'2020. The Big Data Cup challenges involve a released dataset and a well-defined problem with clear evaluation metrics. The challenges run on a data competition platform that maintains a leaderboard for the participants. In the presented case, the data constitute 26336 road images collected from India, Japan, and the Czech Republic to propose methods for automatically detecting road damages in these countries. In total, 121 teams from several countries registered for this competition. The submitted solutions were evaluated using two datasets test1 and test2, comprising 2,631 and 2,664 images. This paper encapsulates the top 12 solutions proposed by these teams. The best performing model utilizes YOLO-based ensemble learning to yield an F1 score of 0.67 on test1 and 0.66 on test2. The paper concludes with a review of the facets that worked well for the presented challenge and those that could be improved in future challenges.

## 1 Introduction

Road maintenance plays a vital role in the socio-economic development of a country. It requires regular assessment of road conditions, which is usually carried out individually by several state agencies. Some agencies perform pavement condition surveys using road survey vehicles equipped with many sensors to evaluate pavement conditions and deterioration. In these vehicles, laser line-scan cameras and three-dimensional (3D) cameras are generally employed to capture pavement surfaces with the best possible quality and resolution. However, such imaging equipment mounted on dedicated vehicles is expensive and is often unaffordable for local agencies with limited budgets. This leads to the requirement of low-cost methods capable of comprehensively surveying road surfaces, such as methods that can be implemented using Smartphones.

Recently, a preliminary version of such a system supporting Smartphone-based Road condition monitoring has been developed by the University of Tokyo, Japan (Maeda et al., 2018). Further, the Japanese research team hosted a challenge (IEEE BigData Cup) to evaluate the contemporary methods working towards the same goal in December 2018. This challenge was the first Road Damage Detection Challenge of the series and was organized in association with the IEEE International Conference on Big Data'2018 held at Seattle, USA. The challenge received wide attention from researchers all over the world (Alfarrarjeh et al., 2018; Kluger et al., 2018; Mandal et al.,2018; Manikandan et al., 2018; Wang et al., 2018a; Wang et al., 2018b). In total, 54 teams participated in the challenge, and several novel methods were proposed for improving the accuracy of automatic road damage detection system.

---

[*] Deeksha@ct.iitr.ac.in

After the road damage detection challenge in 2018, several municipalities in Japan started utilizing the proposed automatic road damage detection system. The practical use and the feedbacks of government agencies suggested that the algorithms need to be more robust. Furthermore, it was observed that most of the existing models are limited to road conditions in a single country. The development of a method that applies to more than one country leads to the possibility of designing a stand-alone system for road damage detection all over the world.

Considering the requirement, Arya et al.(2020) augmented the Japanese dataset with road damage images from India and the Czech. The newly proposed data, Road Damage Dataset – 2020 (RDD-2020), comprises 26620 images, that is almost thrice the volume of the famous 2018 dataset.

The authors carried out a comprehensive analysis of models trained using different combinations of the data.

The RDD-2020 data proposed by Arya et al. has been made publicly available and forms the basis for organizing the second challenge of the road damage detection challenge series, named Global Road Damage Detection Challenge (GRDDC). GRDDC is an online event organized in association with the IEEE International Conference on Big Data'2020. It invites models capable of efficiently detecting road damages for India, Japan, and Czech.

The challenge is intended to open new horizons of possibilities where just the smartphones and drive cameras would be enough for road inspections, not for a single country but all the countries across the world. This paper presents a summary of this challenge, along with the solutions proposed by the participants.

## 2    CHALLENGE REVIEW

### 2.1    Overview

The GRDD challenge is designed to push state of the art in detecting road damages forward. The challenge comprises two components: (i) a publicly available dataset of road images and annotation, and (ii) an online competition and workshop. The GRDDC'2020 dataset consists of annotated road damage images collected from India, Japan, and Czech. There are two key challenges: classification – "does the image contain any instances of a particular road damage class (where the road damage classes include longitudinal cracks, transverse cracks, potholes, etc.)?", and detection – "where are the instances of a particular road damage class in the image (if any)?"

The data challenges are issued with deadlines each year, and a workshop is held to compare and discuss that year's results and methods. The datasets and associated annotation and evaluation methods are subsequently released and available for use at any time. The procedure for GRDDC'2020 is similar to and is inspired by the PASCAL VOC challenge that was organized from 2005 to 2012(see Everingham et al., 2010, and Everingham et al., 2015).

The objectives of the Road Damage Detection challenges are twofold: first to provide challenging road damage images and high-quality annotation, together with a standard evaluation methodology to compare the performance of algorithms fairly (the dataset component); and second to measure state of the art for the current year (the competition component).

This paper aims to describe the challenge, the methods used, the evaluation carried, and the results. Additionally, the state-of-the-art in road damage detection is also covered, as measured by the solutions submitted to the challenge. We focus mainly on the 2020 challenge, GRDDC, as this is the most recent.

### 2.2    Dataset

The data released for GRDDC constitute road images collected from India, Japan, and Czech. The challenge divides the data in three chunks: Train, Test1, andTest2. The set "Train" consists of road images with annotations in XML files in PASCAL VOC format. The other two sets are released without the annotations for evaluating the solutions proposed by the participants. The evaluation procedure is described in the next section.

The statistics for the distribution of images in the three datasets and the three countries are provided in figure 1. The training data contains annotation with marked labels for different road damage types. Figure 2 shows the number of instances for each damage type in training data.

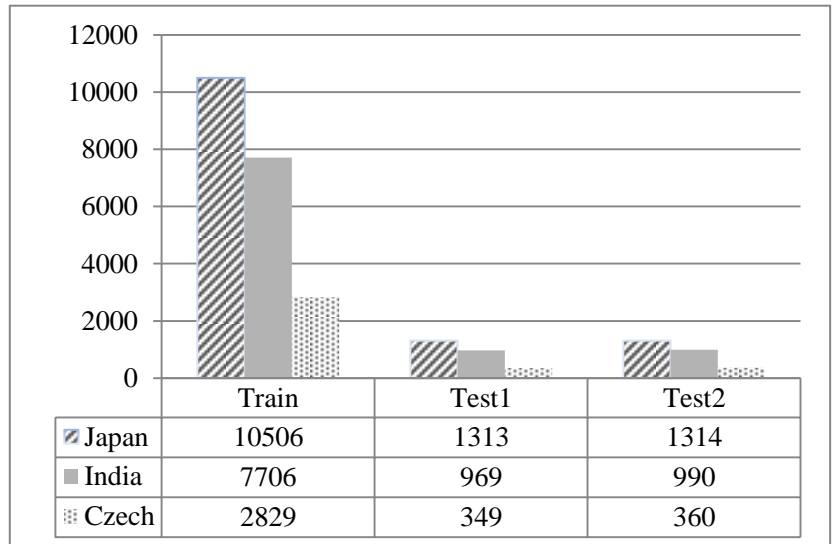

Fig. 1. Dataset Statistics

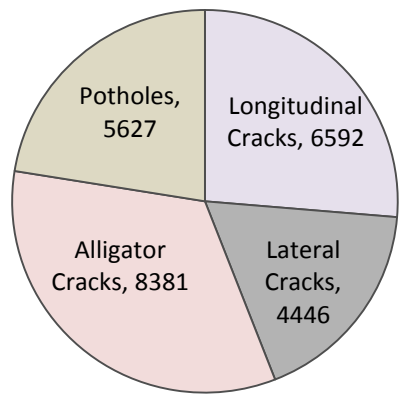

Fig. 2. Distribution of different damage type instances in training data

Our companion paper (Arya et al., 2020) presents the details of the data collection methodology, the damage categories considered, and the corresponding reason, the study area, the annotation procedure, etc. These details are not repeated here.

It is worth noting that although the data proposed by Arya et al.(2020) is utilized for organizing the competition, the data statistics are slightly different. This is because some of the images used by Arya et al.(2020) have not been included in the challenge data due to privacy reasons.

### 2.3 Tasks

GRDDC required the participants to propose an algorithm that can automatically recognize the road damages present in an image captured from any of the following three countries: India, Japan, and

Czech. The recognition implies detecting the damage location in the image and identifying the damage type.

Using this algorithm, the participants need to predict labels and the corresponding boundary boxes for damage instances present in each test data image. The predictions need to be submitted as a CSV file through the submission link provided on the website. The general rules to be followed while accomplishing the challenge tasks are stated below:
  (i)   Algorithm: Recently, Deep Learning models have become the first choice for applications involving object detection. However, the challenge allows all types of algorithms as long as the proposed solution can outperform other solutions.
  (ii)  Pre-Trained Models: The challenge permits the use of Transfer learning utilizing pre-trained models.
  (iii) Data: Participants are restricted to train their algorithms on the provided Road Data train sets. Collecting additional data for the target attribute labels is not allowed. However, it is permitted to increase training images using data augmentation, GANs, etc., artificially.
  (iv)  License: The dataset released for the challenge is publicly available under the Creative Commons Attribution-ShareAlike 4.0 International License (CC BY-SA 4.0).

The following section provides details about the submission and evaluation of the results.

## 3    SUBMISSION AND EVALUATION

### 3.1    Submission of Results

The running of the challenge consisted of four phases with submission at multiple stages, as described below.
  (i)   At the start of the challenge, the task was announced, and training/validation images with annotation (stored in an XML format compatible with LabelImg tool) were made available to the participants.
  (ii)  In the second phase, unannotated test images from the test1 set were distributed. Participants were then required to run their methods on this data and submit results to an evaluation server. The server is designed with an evaluation script, which calculates a score corresponding to every submission. The calculated score is added to the private leaderboard of the participants. Simultaneously, a public leaderboard is maintained on the website to report the highest score achieved by the teams on their private leaderboard. The test1 data was available for approximately four months before submitting results to give ample time to participants for training their models.
  (iii) In the third phase, another test dataset, test2, comprising 2,664 images, was released. This dataset was released just 15 days before the submission deadline. Similar to the test1, a new public leaderboard was created for the test2. Additionally, each team was assigned its private leaderboard for performing the experiments on test2.
  (iv)  In the final phase, the participants were required to submit their source code to verify the replicability of the proposed algorithms.

The evaluation procedure for the submissions is described in the following sub-section.

### 3.2    Evaluation

The GRDD challenge evaluates the results submitted by the participants using an evaluation script embedded in the server. The script utilizes two inputs: (i) the ground truth information of the test dataset, the annotations, and (ii) the file containing the participants' predictions. A prediction is considered correct if it satisfies the following two criteria:

(i) The area of overlap between the predicted bounding box and ground truth bounding box exceeds 50%. The condition is described as IoU > 0.5. IoU denotes Intersection Over Union. For more details regarding this, please refer to Arya et al.(2020).
(ii) The predicted label matches the actual label, as specified in the annotation file (ground truth) of the image.

The script compares the two input files and calculates F1-Score for the submission. The F1-score is calculated as the Harmonic Mean of Precision and Recall values. Precision is the ratio of true positives to all predicted positive. The recall is the ratio of true positives to all actual positives. The details of the parameters are given below:

(i) True Positive (TP): When a damage instance is present in the ground truth, and the label and the bounding box of the instance are correctly predicted with IoU > 0.5.
(ii) False Positive (FP): When the model predicts a damage instance at a particular location in the image, but the instance is not present in the ground truth for the image. This also covers the case when the predicted label doesn't match with the actual label.
(iii) False Negative (FN): When a damage instance is present in the ground truth, but the model fails to predict either the correct label or the bounding box of the instance.
(iv) Recall

$$Recall = \frac{TP}{TP + FN} \quad (1)$$

(v) Precision

$$Precision = \frac{TP}{TP + FP} \quad (2)$$

(vi) F1 − Score

$$F1 - Score = 2\frac{Precision.Recall}{Precision + Recall} \quad (3)$$

The F1 metric weights recall and precision equally. Thus, the competition favors moderately good performance on both over the outstanding performance on one and poor performance on the other. F1 is calculated separately for test1 and test2. Finally, the average of both the scores is used to rank the teams.

## 4 Global Road Damage Detection Challenge'2020: Results and Rankings

In total, 121 teams from all over the world registered for the challenge. The related data is available on the challenge website. 13 teams were shortlisted based on the average F1-score achieved by the proposed solutions and the submissions od source-code. Out of these, 12 teams made it to the final round requiring a detailed report of their proposed solution (Doshi and Yilmaz, 2020; Mandal et al., 2020; Pham et al., 2020; Tristan et al., 2020). Table 1 lists the ranking achieved by these teams and their scores corresponding to the two leaderboards.

The details of the contributors and the shortlisted teams' methods are presented in tables 2 and 3, respectively. The top 3 teams achieved an average F1-score of more than 60% and proposed efficient methods using ensemble learning and several data augmentation techniques.

The winning method was the one proposed by team IMSC (Hedge et al., 2020) that was based on ultralytics-YOLO (u-YOLO) [YOLOv5, 2020]. The proposed approach applied the test time augmentation (TTA) procedure on test data to improve their model's robustness. TTA augments the data by using several transformations (e.g., horizontal flipping, increasing image resolution) on each test image and generating new images.

Table 1: GRDDC Ranks and Scores of top 12 teams

| GRDDC – Rank | Name of the Team | Test1-Score | Test2-Score | Link for the Source Code |
|---|---|---|---|---|
| 1 | IMSC | 0.6748 | 0.6662 | https://github.com/USC-InfoLab/rddc2020 |
| 2 | SIS Lab | 0.6275 | 0.6358 | https://github.com/kevaldoshi17/IEEE-Big-Data-2020 |
| 3 | DD-VISION | 0.629 | 0.6219 | https://pan.baidu.com/s/1VjLuNBVJGS34mMMpDkDRGQ  password: xzc6. |
| 4 | titan_mu | 0.5814 | 0.5751 | https://github.com/titanmu/RoadCrackDetection |
| 5 | Dongjuns | 0.5683 | 0.5710 | https://github.com/dongjuns/RoadDamageDetector |
| 6 | SUTPC | 0.5636 | 0.5707 | https://github.com/ZhangXG001/RoadDamgeDetection |
| 7 | RICS | 0.565 | 0.547 | https://github.com/mahdi65/roadDamageDetection2020 |
| 8 | AIRS-CSR | 0.554 | 0.541 | https://github.com/ZhangXG001/RoadDamgeDetection |
| 9 | CS17 | 0.5413 | 0.5430 | https://github.com/TristHas/road |
| 10 | BDASL | 0.5368 | 0.5426 | https://github.com/vishwakarmarhl/rdd2020 |
| 11 | IDVL | 0.51 | 0.514 | https://github.com/iDataVisualizationLab/roaddamagedetector |
| 12 | E-LAB | 0.4720 | 0.4656 | https://github.com/MagischeMiesmuschel/E-LAB_IEEE_BDC_GRDD_2020_submission |

These new images are fed to the trained u-YOLO model along with the existing images. Thus, corresponding to every test image, multiple predictions are generated using the augmented images. Duplicate or overlapped predictions generated in the process are filtered using the non-maximum suppression (NMS) algorithm. The whole approach is referred to as Ensemble Prediction (EP).

Along with EP, the team proposed another approach termed the Ensemble model (EM). The approach EM, as the name suggests, ensembles different variants of u-YOLO models. Given that training a u-YOLO model involves tuning different hyperparameters, using different combinations of these parameters generates different trained models. A subset of these models is selected such that they maximize the overall accuracy. Each image is passed through all the selected models, and predictions from each model are averaged before applying non-maximum suppression. This ensemble technique helps in achieving better accuracy by reducing the prediction variance.

After that, the team combines the two approaches and proposes the final solution termed as Ensemble Model with Ensemble Prediction (EM+EP). In this approach, EM is extended with the TTA procedure used in EP. That is, after transforming a test image using TTA, the augmented images are fed into each model of EM. Then, the predicted bounding boxes from the augmented images for each model are averaged before applying NMS.

The authors compare the performance in terms of speed as well as accuracy for all three approaches (EM, EP, and EM+EP). The statistics show that while the accuracy is improved in the case of (EM+EP) providing the highest F1-score (0.67 for test1), the approach is worst in terms of speed of detection, as measured using Detection time per image. More details are provided in (Hedge et al., 2020).

Table 2: Details of the contributors (Top 12 teams for GRDDC'2020)

| Team Name | Contributors | Affiliation | Country |
|---|---|---|---|
| IMSC | Vinuta Hedge, Dweep Trivedi, Abdullah Alfarrarjeh, Aditi Deepak, Seon Ho Kim, Cyrus Shahabi | The University of Southern California, German Jordanian University | United States of America, Jordan |
| SIS Lab | Keval Doshi, Yasin Yilmaz | University of South Florida | United States |
| DD-VISION | Zixiang Pei, Xiubao Zhang, Rongheng Lin, Haifeng Shen, Jian Tang, and Yi Yang | DD-VISION | China |
| titan_mu | Vishal Mandal, Yaw Adu-Gyamfi, Abdul Rashid Mussah | University of Missouri-Columbia | United States of America |
| Dongjuns | Dongjun Jeong | Robotics Lab, University of Southern Denmark | Denmark |
| SUTPC | Yuming Liu, Xiaoyong Zhang, Bingzhen Zhang, and Zhenwu Chen | Shenzhen Urban Transport Planning Center | China |
| RICS | Sadra Naddaf-Sh, M-Mahdi Naddaf-Sh, Amir R. Kashani, Hassan Zargarzadeh | Lamar University, Stanley Black and Decker | United States of America |
| AIRS-CSR | Xiaoguang Zhang, Xuan Xia, Nan Li, Ma Lin, Junlin Song, Ning Ding | Shenzhen Institute of Artificial Intelligence and Robotics for Society, and Institute of Robotics and Intelligent Manufacturing, The Chinese University of Hong Kong, Shenzhen | China |
| CS17 | Tristan Hascoet, Yihao Zhang, Andreas Persch, Ryoichi Takashima, Tetsuya Takiguchi, and Yasuo Ariki | Kobe University | Japan |
| BDASL | Rahul Vishwakarma, Ravigopal Vennelakanti | Hitachi America Ltd. | United States |
| IDVL | Vung Pham, Chau Pham, Tommy Dang | Texas Tech University | United States of America |
| E-LAB | Felix Kortmann, Kevin Talits, Pascal Fassmeyer, Alexander Warnecke, Nicolas Meier, Jens Heger, Paul Drews, Burkhardt Funk | HELLA GmbH & Co. KGaA | Germany |

Table 3: Details of the Proposed Solutions by top 12 teams for GRDDC'2020.

| Team Name | Proposed Solution | Data Augmentation | Other methods explored |
|---|---|---|---|
| IMSC | Ensemble Learning with Ultralytics-YOLO and Test Time Augmentation | Yes, Synthesized images using Python Augmenter (sharpen, multiply, additive Gaussian noise, and affine texture mapping). | Faster-RCNN, YOLOv3 |
| SIS Lab | Ensemble model with YOLO-v4 as base model. | Yes, A random crop augmentation algorithm that outputs three images for each training image. | YOLO-v4 with different input image resolution |
| DD-VISION | A Consistency Filtering Mechanism and model ensemble with cascade R-CNN as the base model | Yes, road segmentation, mixup, Contrast Limited Adaptive Histogram Equalization(CLAHE), RGB shift | Faster-RCNN, ResNeXt-101, HR-Net, CBNet, ResNet-50 |
| titan_mu | YOLO model trained on CSPDarknet53 backbone | Not used | CenterNet and EfficientDet models trained on Hourglass-104 and EfficientNet backbones |
| Dongjuns | YOLOv5x | Yes, image HSV, image translation, image scale, horizontal flip, mosaic | Test-Time Augmentation with Ensemble models |
| SUTPC | Ensemble(YOLO-v4 + Faster-RCNN) | Yes, deeplabv3+ model trained from citysacapes dataset to build a road-interest map that contains the road and the front of car information. | YOLOv4 and Faster-RCNN with different hyper-parameter settings |
| RICS | EfficientDet | Explored the auto augmentation policies for object detection but did not use in the final solution. | Auto augmentation policies for object detection |
| AIRS-CSR | YOLOv4 | Yes. Image translation by adjusting brightness, contrast, hue, saturation, and noise. The random scaling and Mosaic data augmentation, Conditional GAN | GAN, Mosaic data augmentation, random scaling, Optimized Anchors |
| CS17 | Resnet-18 and Resnet-50 backbones based Faster-RCNN two-stage detection architecture | Tried, but that didn't improve the accuracy | Data augmentation, Ensemble and Test Time augmentation |
| BDASL | Multi-stage Faster R-CNN with Resnet-50 and Resnet-101 backbones | Yes, attempted to resize and semantic segmentation of the road surface. | Yolov5 with CSPNet backbone |
| IDVL | Road Damage Detector using Detectron2 and Faster R-CNN | Yes, horizontal flipping and resizing. | Data augmentation by adjusting brightness, contrast, cropping, generating artificial damages, etc. |
| E-LAB | FR-CNN; Classifying the region and using regional experts for the detection | Random horizontal flip, random adjustment of hue, contrast, saturation, and random cropping | Tested if Homogeneous Road damage distribution could help in improving the results. |

# 5 Discussion and Future Scope

The significance of Road Damage Detection challenges lies in providing a rich dataset and a standardised evaluation framework for comparing different methods. The participation has increased rapidly since the challenge was first introduced (54 teams in 2018 to 121 teams in 2020). The use of the associated dataset and the leaderboards too has shown the same popularity. After the challenge submissions were closed and the leaderboards were freezed, we received several requests to re-open the leaderboards. To address these requests while maintaining the challenge timeline, new leaderboards have been created on the challenge website. The researchers have been permitted to perform experiments and submit their solutions to leaderboards for both test1 and test2 even though the new submissions would no longer count for the 2020 competition.

## 5.1 The Positive Points of the challenge

The tremendous progress, success and popularity of the road damage detection challenge can be attributed to the following factors:

(i) *Standard Method of Assessment*: The challenge provides a standard method of assessment where the participants can submit their output file and the score would be calculated by the software deployed on the challenge website automatically. In the absence of such standard method of assessment, the authors of different papers may evaluate the performance of their models in different ways leadig to results that can not be compared directly. For example some authors may give equal weight to each image and others may give equal weight to each class.

(ii) *Fixing the test data for evaluating the performance*: The challenge released the dataset in separate train and test sets. This helped in avoiding the cases where different teams may have chosen different images for testing purpose. Since the data contains multiple classes with imbalance number of instances, the performance for different classes vary significantly. For example, a team choosing test images with instances of alligator cracks in majority would have achieved better score than the team that by chance chose the test images with more transverse crack instances even if the same model is used. Arya et al.(2020) elaborate the reason for poor performance in detecting transverse cracks. Similarly, a team choosing more images from Japan for testing purpose would have an advantage over the team using images from the Czech Republic. This is because the ratio of Japanese images is significantly higher than that of images collected from the Czech Republic. Thus, fixing the test datasets accounted for better comparability of the solutions proposed by different teams and authors.

(iii) *Augmenting the data gradually*: The road damage data introduced in 2018 challenge included 9053 road images collected from Japan. Generating new dataset for road damage is a labourious and time taking process as it requires annotating each image for all type of damages that have been considered. Consequently, Maeda et al.(2020) augmented the dataset by using Generative Adverserial Network, increasing the number of images to 13135. For 2020 challenge, GRDDC, the data was augmented using new images collected from India and the Czech Republic. This helped in increasing the size as well as the coverage area of the roads. Further, inclusion of data from different countries helped in collecting images of different road textures with different degree of road damage.

Despite the overwhelming response received this year, there are several directions in which the dataset and the challenge can be improved to retain its popularity and meet the requirements of research community in future. The next segment discusses the same.

## 5.2 Scope for Future and the room for improvement

Following are some of the directions that can be considered for the future road damage detection challenges.

(i) Currently, the RDD challenges consider solutions based on the methods trained on only the provided training data. In the future, methods trained on any other data may also be allowed by defining a new category for the competition, similar to the famous PASCAL VOC challenges.

(ii) Until now, RDD challenges focussed only on the detection and classification of road damages. Future challenges may introduce new tasks such as Severity analysis, Pixel-level damage analysis for covering a wider spectrum for road condition monitoring.

(iii) Further, multiple evaluation metrics can be introduced for assessing the performance better. For instance, considering the inference speed and size of the proposed model in addition to F1-Score would be useful for developing the real-time smartphone-based object detection methods.

(iv) Additionally, instead of using just the average F1-score, future challenges may also consider the models' performance for individual damage classes.

(v) Furthermore, for the upcoming years, the road damage dataset could be augmented to include a more balanced representation of damage classes.

## 6 Conclusion

The Global Road Damage Detection challenge focuses on the detection and classification of road damage in different parts of the world, particularly India, Japan, and the Czech Republic. It includes four common damage categories: Longitudinal Cracks, Transverse Cracks, Alligator Cracks, and Potholes. Since the data comes from different regions of the world, including developed and developing countries, the quality of road hardening varies. Further, the degree of damage, area, aspect ratio, etc., of each type of damage differs significantly.

The current challenge was a huge success, with 121 registered teams. This paper summarizes the challenge and provides valuable insights for three audiences: *algorithm designers*, researchers who are interested in the latest algorithms for object detection applications; *challenge designers,* who are interested in the step-wise organization of the challenge*; and transport engineers* interested in solutions for automating the detection of road damages.

Besides, the paper provides inputs for designing future challenges, such as including additional tasks (crack ratio analysis, potholes severity analysis, road segmentation, etc.), and evaluation metrics for more efficient automatic monitoring of road conditions.

## 7 Acknowledgments

We thank all the participants of the challenge for their contributions. We also thank the sponsor UrbanX Technologies for providing the requisite funds. Additionally, we acknowledge the support received from Sekimoto-Lab, Institute of Industrial Science, The University of Tokyo, Japan, and IIT Roorkee, India, for organizing the challenge successfully.